\documentclass[runningheads]{llncs}

\usepackage{eccv}

\usepackage{eccvabbrv}

\usepackage{graphicx}
\usepackage{booktabs}
\usepackage{times}
\usepackage{epsfig}
\usepackage{amsmath}
\usepackage{amssymb}
\usepackage{multirow}
\usepackage[table]{xcolor}

\usepackage[accsupp]{axessibility}  

\usepackage{hyperref}

\usepackage{orcidlink}

\begin{document}

\title{Explicit Over Implicit: Enhancing CNNs Via Complex Structure Tensor Representations for Periocular Recognition} 


\author{Kevin Hernandez-Diaz\inst{1}\orcidlink{0000-0002-9696-7843} \and
Josef Bigun\inst{1}\orcidlink{0000-0002-4929-1262} \and
Fernando Alonso-Fernandez\inst{1}\orcidlink{0000-0002-1400-346X}}

\authorrunning{K. Hernandez-Diaz et al.}

\institute{Halmstad University, Sweden \\
\email{\{kevin.hernandez-diaz, josef.bigun, fernando.alonso-fernandez\}@hh.se}}

\maketitle

\begin{abstract}
  Our study provides evidence that CNNs struggle to extract orientation features effectively. We show that using the Complex Structure Tensor, which contains compact orientation features with certainties, as input to CNNs consistently improves identification accuracy compared to grayscale inputs alone. Experiments also demonstrated that our inputs, provided by mini-complex conv-nets, combined with reduced CNN sizes, outperformed full-fledged, prevailing CNN architectures. This suggests that the upfront use of orientation features in CNNs, a strategy seen in mammalian vision, not only mitigates their limitations but also enhances their explainability and relevance to thin-clients. Experiments were conducted on publicly available datasets comprising periocular images (Cross-Eyed and PolyU) for biometric identification and verification in both Close-World and Open-World Scenarios using six CNN architectures. Our experiments on the Cross-Eyed and PolyU datasets yield a 5-26\% reduction in EER, providing strong empirical evidence that explicit orientation priors mitigate CNN representational limits in Open-World and Close-World scenarios.
\end{abstract}

\section{Introduction}
\label{intro}

Computer Vision (CV) defines texture as a repetitive pattern \cite{tuceryan1993texture}. Extracting texture features amounts to finding a vector of parameters from subsets of the pattern to uniquely characterize it. Ideally, the vector should be invariant to translation within a given texture and different across textures. 
The squared magnitude of the Fourier Transform, FT, a.k.a. the power spectrum, of an image is invariant to image translations.  Consequently, the power spectrum of local images has been a main source of texture perception \cite{julesztexton}, and texture feature vectors \cite{daugman83}\cite{HOG}\cite{tuceryan1993texture}\cite{LBP}. Even Convolutional Neural Networks (CNN) are biased towards texture information \cite{geirhos2018imagenettrained}, showing that first-layer ImageNet-trained CNNs' filters have Gabor-like shape \cite{krizhevsky2012imagenet}. Since Hubel and Wiesel’s Nobel Prize-winning work \cite{hubel59}, we know that mammalian vision is enabled by cells, each tuned to a unique orientation of translating bars in its visual field, and are upfront in the brain’s data flow, at the visual cortex. Some cells are invariant to translation, whereas tune-in angles are uniform, separated by less than $ 1^\circ$ \cite{schwartz}. Gabor function \cite{gabor} magnitude responses approximate this processing \cite{kulikowski81}.


Texture plays a crucial role in many image-based biometric recognition systems, such as the case of fingerprints \cite{jain2001fingerprint}, face \cite{gaborface1}, iris \cite{daugman2007new}\cite{daugman2009iris}, and palm vein \cite{mirmohamadsadeghi2014palm}. In 2009, the authors of \cite{park2009periocular} showed the potential of the periocular region for biometric recognition, which they defined as the facial area in the immediate vicinity of the eye. This region encompasses the iris, eyebrows, eyelids, commissures, skin texture, and the general shape of the eye, which can be used for recognition. This information-rich area has not only shown to be quite stable but also to achieve high performance in recognizing people's gender, ethnicity, and identity \cite{alonso2016survey}. The periocular area provides flexibility regarding acquisition and occlusion as a middle ground between face and iris recognition. Like its biometric siblings, it has been the target of several texture-extraction algorithms for recognition with good performance \cite{periocularlbp}\cite{hernandez2023one}.

In this paper, we propose a combination of a texture-extraction algorithm based on up to the Second Order Complex Derivative of Gaussians \cite{[Bigun04]} to detect the presence of linear symmetry patterns in a region to enhance the performance of CNNs. The main contributions of this paper are as follows:
\begin{itemize}
    \item We frame our central hypothesis: CNN networks alone struggle to produce orientation features significant for decision-making with training data of reasonable size ($\approx$ ten thousand images). 
     \item We show that CNN networks can benefit from compact orientation features as texture features at the input by improving their a) explainability and focus, b) convergence speed with small training data, c) network size, and d) performance.
     \item We suggest using complex STs as a mini-convnet, relying on complex scalar products and complex non-linearities to provide compact inputs to CNNs.
     
 \end{itemize}


\section{Related Work}\label{background}

Several successful traditional Computer Vision algorithms have focused on texture extraction. This section summarizes some algorithms that follow a similar approach to ours by combining them with CNNs.

This paper combines CNNs with a compact texture-orientation descriptor based on Structure Tensor Theory \cite{[Bigun04]}. As detailed in Section \ref{LST}, it uses complex derivatives of Gaussians to extract gradients and estimate the presence and orientation of line patterns. While symmetry descriptors are well known in biometrics for their robustness to viewpoint and scale changes \cite{ref}, their contribution in deep learning has been largely unexplored.

Prior work has applied related symmetry-based filters to tasks such as fingerprint core detection \cite{[Bigun04]}, eye detection via concentric-circle patterns \cite{GSTeyedetection}, and iris segmentation using circle-detecting complex filters \cite{ref1}, achieving state-of-the-art results \cite{ref2}. Extended symmetry families have also been used for facial landmark detection and hand-motion analysis \cite{ref0}\cite{ref4}. For recognition, several studies leveraged multi-scale symmetry responses around keypoints for periocular and forensic fingerprint identification, reaching near state-of-the-art performance \cite{symmetryfingerprint}.

To leverage complementary strengths, many works combine handcrafted descriptors with CNNs, either via multi-stream designs or by embedding descriptors into the network. \cite{LBPCNNfeatures} combined LBP and RGB features with CNNs for face anti-spoofing, increasing performance while reducing network size. \cite{periocularlbp} proposed a double stream CNN that takes RGB ocular images and LBP features, improving recognition. \cite{LBPNet} replaced standard convolutions with Local Binary Convolution layers, matching CNN performance while cutting learnable parameters by 9–169 times.

HOG has also been combined with CNNs to boost performance. For example, \cite{HOGCNNtracking} fused HOG with CNN features for efficient multi-camera pedestrian tracking and re-identification. In periocular recognition, \cite{hogcnnperiocular} combined handcrafted HOG, pre-trained CNN features, and gender cues from a shallow network. For video-based facial expression recognition, \cite{hogcnnexpression} applied HOG on shallow spatial and temporal CNN features.

Several works have combined Gabor filters with CNNs. In \cite{circulargaborcnn}, magnitude responses from complex local circular Gabor filters are used to augment iris data for training a multi-scale feature-fusion network for recognition. GaborNet \cite{gabornet} constrained first-layer filters to the Gabor function by learning Gabor parameters rather than raw weights, improving accuracy while reducing learnable parameters. \cite{gaborensemble} applied a Gabor filter bank to form multiple Gabor responses and trained an ensemble of CNNs, each specialized to a specific response for face recognition. Gabor-based CNN inputs have also improved facial expression and soft-biometric recognition: \cite{gaboremotions} used outputs from two successive Gabor filters to train a CNN for emotion recognition, reporting better performance and faster convergence, while \cite{gaborcapsule} showed Gabor responses boost accuracy for both standard CNNs and capsule networks on face-based soft biometrics.


Prior work has also combined multiple handcrafted descriptors to complement CNNs. For instance, \cite{alonso2020cross} fused scores from CNNs and several traditional features (e.g., LBP, HOG, and symmetry descriptors), improving cross-sensor periocular recognition.



\begin{figure}[h!]
\centering
\includegraphics[width=0.90\columnwidth]{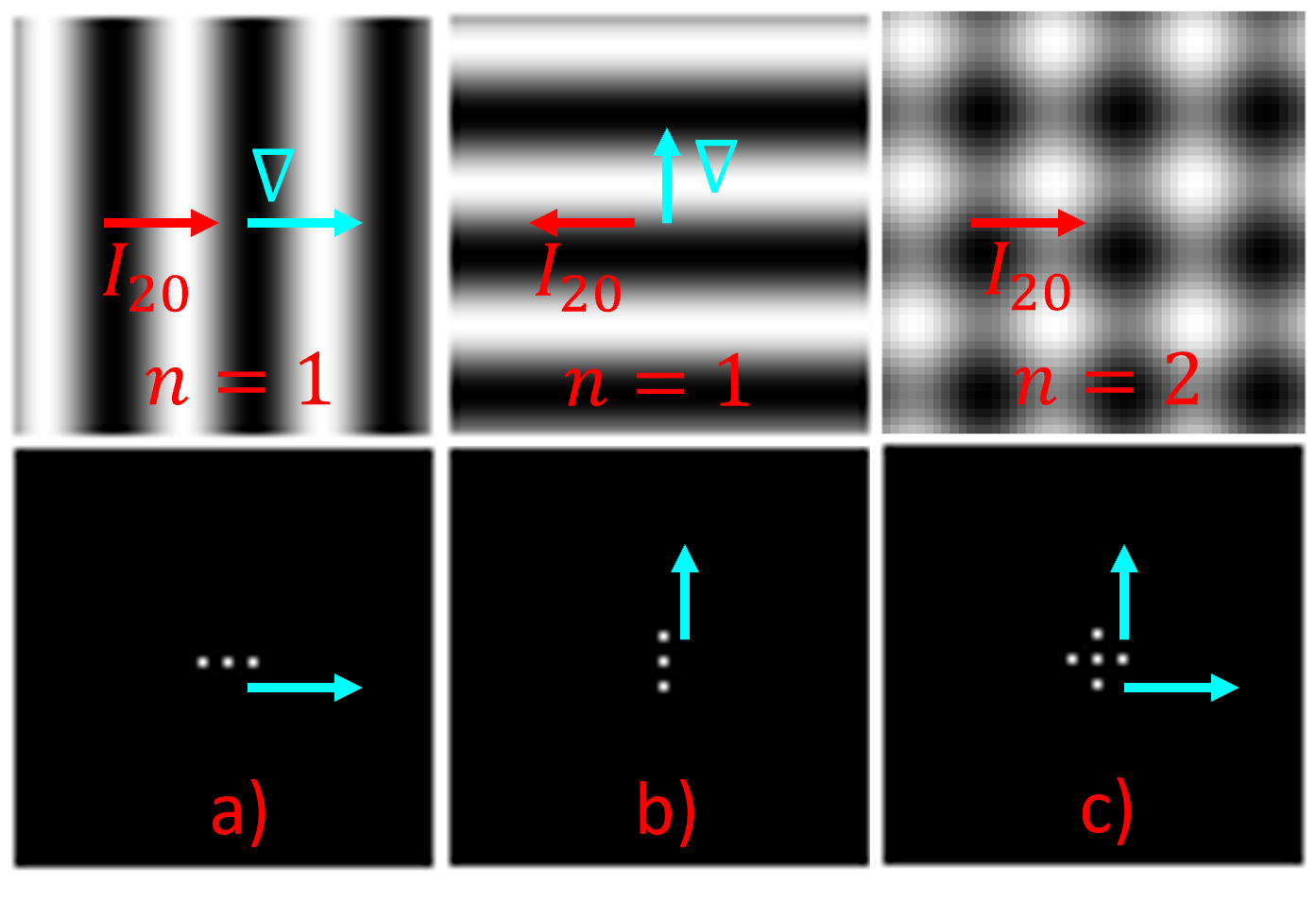}
\caption{a),b): Examples of  1-folded (linearly) symmetric textures of planar waves with $ I_{20}$ $  
n=1$  and gradient vectors; c) dito but  2-folded symmetric, with $ n=2$. FT magnitudes are below.  \label{fourier}}
\end{figure}

\begin{figure*}[ht]
\centering
\includegraphics[width=0.90\textwidth]{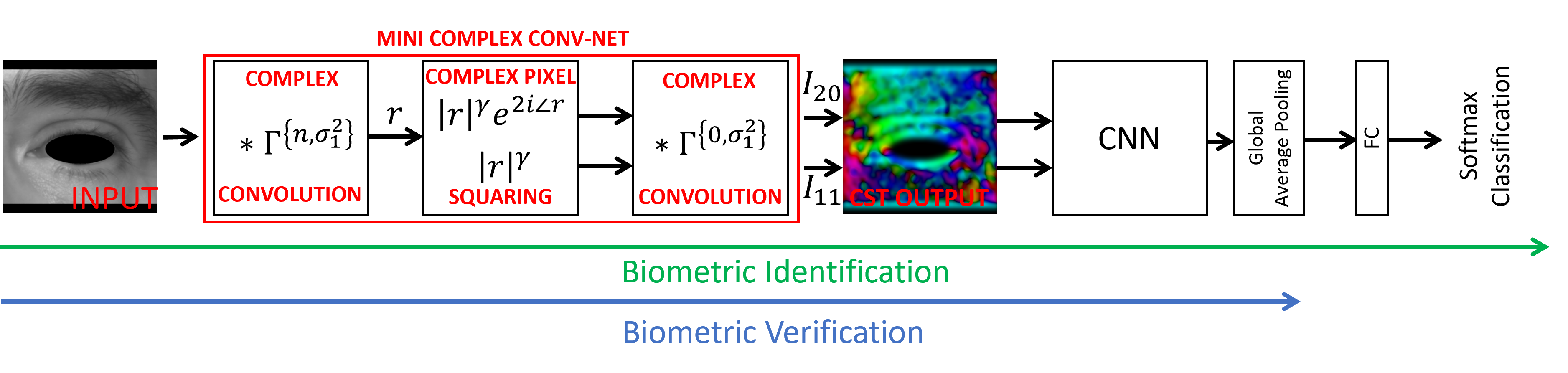}
\caption{Pipeline of the proposed method. \label{I20process}}
\end{figure*}

\section{Structure Tensor  and FT concentrations}
\label{LST}


We study the 2D Structure Tensor (ST) theory to quantify line-pattern texture in a local image neighborhood. Following \cite{bigun87london}, ST detects symmetries in the neighborhood’s power spectrum that correspond to translation-invariant line orientations.
ST corresponds to the second-order angular moments of the local power spectrum \cite{bigun87london}, i.e. those of the magnitudes of Gabor responses,  providing interpolation of tune-in orientations. 

We first summarize the simplest pattern class: linearly symmetric patterns, in which all iso-gray curves in the neighborhood share a single direction.
This class includes not only lines and edges, but also parallel line patterns and planar waves of the form $A\exp (i\omega_x x
+i\omega_yy) $ are in the family (see Fig. \ref{fourier} a), b)).

ST is the $ 2\times2$ symmetric real semipositive definite matrix obtained by averaging tensor products of
gradients in a neighborhood image $ f $, $ \mathbf {ST} =\langle\nabla \! f \nabla^T\!\! f\rangle$. 
It is the second-order real moment matrix of the power spectrum, scatter, or inertia, used for fitting an optimal axis in mechanics. The matrix $ \mathbf {ST} $ encodes the presence of a linearly symmetric pattern family in the neighborhood and is determined via 3 real variables, e.g., non-redundant elements of the $ \mathbf{ST}$.
Satisfying $\mathbf {ST}  u=\lambda u $, eigenvalues $0\le \lambda_2\le \lambda_1$ with corresponding (orthogonal) eigenvectors $ u_2$ and $ u_1 $ are useful to explain the neighborhood in indirect human interpretation, in terms of physical observation (orientation).

The matrix $\mathbf {ST} $ can be decomposed as 
\begin{equation}
\mathbf{ST} =(\lambda_1-\lambda_2)u_1u_1^T+ \lambda_2
(u_1u_1^T+u_2u_2^T) \label{eq:spectral} 
\end{equation} 
where $ u_1u_1^T+u_2u_2^T=E$ is the unity matrix hinting at $ \mathbf {ST}$  
can be uniquely decomposed into an oriented "line" component (encoded by $u_1u_1^T$) with weight-power of $ \lambda_1-\lambda_2$ and a non-oriented,  "balanced", component (encoded by E) with the weight-power of $\lambda_2$. 

The total available weight-power of all directions (=eigenvectors) is $\lambda_1+\lambda_2 $. When the neighborhood is linearly symmetric, we will have $\lambda_2=0$. Also, the representation of the neighborhood direction by $u_1u_1^T $, rather than $u_1 $, encodes continuously the fact that a neighborhood belonging to the family is invariant to rotation by $\pi$ radians since $(-u_1)(-u_1)^T=u_1u_1^T$. Such patterns possess then 2-folded rotational symmetry, $ 2\pi/2$, in the neighborhood and in the spectrum. Equally important is that $ \mathbf{ST}$ itself is invariant to translation so long as the neighborhood remains inside the texture (of the linearly symmetric pattern). 

The matrix $\mathbf{ST}$ represents the continuous total least square error solution of fitting an axis to the power spectrum of the neighborhood. However, if and only if the neighborhood is linearly symmetric, its entire spectral power collapses to 2 or more Dirac distributions located on a line through the (spectral) origin. In the solution, the error of the best fit is $ \lambda_2$ (ideally zero), and the worst fit is $\lambda_1$. Accordingly, the difference  $\lambda_1-\lambda_2 $ is "how much better" the best error is w.r.t. the worst, representing the confidence in the fit, whereas $\theta=\angle u_1$, is the direction of the pattern. However, the confidence is energy-dependent and can vary with the neighborhood's contrast.

To exploit the ST theory more effectively, complex numbers can be used to represent 2D vectors, including gradients. The main advantage of this "overhaul" is the mathematical completeness of STs via power series expansions, allowing the representation of $ 2n$-folded symmetries with increasing $n$. First, we exploit the second-order complex moments of the neighborhood power spectrum, $I_{20}$ and $I_{11}$.    

\begin{equation}
\label{LST_eq}
    \mathbf{CST} \!= \!\begin{pmatrix}
I_{20},
 I_{11}
\end{pmatrix} \!=\! 
\begin{pmatrix}
(\lambda_1 \! - \!\lambda_2)\exp({2i\angle u_1 }) ,\,
\lambda_1 \!+ \! \lambda_2
\end{pmatrix} 
\end{equation}

where $ i=\sqrt {-1} $, and the vector has 3 degrees of freedom. Components
are readily connected to  $\lambda_1,\lambda_2, u_1$ of $\mathbf {ST}$, (\ref{eq:spectral}). The CST vector is thus the complex equivalent of ST \cite{bigun87london}. The vector has the same degree of freedom as the symmetric $ \mathbf{ST}$, 3, since $I_{20}$ is complex and $I_{11} $ is real and non-negative. 
Replacing the tensor mapping $ u_1 u_1^T$ in 2D, the mapping $2\angle u_1$ is similarly invariant to pattern rotation with $ \pi $ i.e. $2(\angle u_1+\pi)=2\angle u_1$.

Because of $0\le\lambda_2\le \lambda_1$ and the principle of collapse, the $\lambda_1+\lambda_2$ is a tight upper bound for the confidence $0\le \lambda_1-\lambda_2 \le  \lambda_1+\lambda_2$ as $\lambda_2$ approaches zero. An exploitable fact is then, that confidence equals the upper bound if and only if the pattern is linearly symmetric.


We can estimate these moments without performing an FT, Gabor filtering, or computing the eigenvalues of $\mathbf {ST} $. As illustrated by  Fig. \ref{I20process}, they can be obtained directly via a mini complex convolutional network, comprising a convolution of the original image with complex filters generated by 
\begin{equation}
\label{eq:gammader} 
\Gamma^{ \{ n, \sigma^2\} }(x,y)= (D_x+iD_y)^n G_{\sigma^2} (x,y)\} 
\end{equation}
where, 
\begin{equation}
G_{\sigma^2} (x,y)=\frac {1} {2\pi \sigma^2}
\exp(- \frac {x^2+y^2} {2\sigma^2} ), 
\end{equation} 
is the familiar Gaussian function with variance $\sigma^2 $. 

The {\bf linear symmetry algorithm} generates elements of CST, as follows. 
\begin{enumerate}
 \item Convolve gray image $f$ with the filter $\Gamma^{\{n,\sigma^2\}}$, by setting $n=1$, and $\sigma=\sigma_1$. The result is a complex-valued image, $r$.  
 \item Square each pixel in $ r $ while emphasizing (gradient) magnitudes with $ \gamma$ to yield two  outputs, complex $r_a=|r|^\gamma \exp(2\angle r) $ having double gradient angles, and real $ r_b=|r|^\gamma$.
 \item[3a] Convolve the $r_a $ strand of Step 2 with the Gaussian filter $ \Gamma^{\{n,\sigma^2\} }$ where $n=0$, and $\sigma=\sigma_2$ to provide the complex $I_{20}$. 
 \item[3b] Convolve the $r_b $ strand of Step 2 with the same Gaussian filter to output the real $I_{11}$.
\end{enumerate}

Linearly symmetric textures have their 2D power spectra concentrated into a line, as illustrated in Fig. \ref{fourier} a), b). 
However, for more complex patterns with multiple orientations, as Fig. \ref{fourier} c) we need to include higher-order complex moments \cite{mikaelyan14incheon}, i.e. $ n>1$. By increasing  $n$ in the first step, the algorithm instead computes elements of the vector ($ I_{2n,0} $, $I_{n,n}  $). Such increases yield even orders of complex moments of the power spectrum, to check for its {\em $ 2n$-folded symmetry} concentration. Instead of 1 axis as in $ n=1$,  each $(I_{2n,0}, I_{n,n} )$ pair fits $n$ evenly distributed axes through the origin, i.e. multiple axes having  $ 2n-$folded symmetry.



\begin{table*}[ht!]
\centering
\resizebox{0.90\textwidth}{!}{%
\begin{tabular}{|c|c|c|c|c|c|c|}
\hline
\textbf{Database} &
  \textbf{Protocol} &
  \textbf{Total images (classes)} &
  \textbf{Train Part. Images (Classes)} &
  \textbf{Gen./Imp. Pairs} &
  \textbf{Test Part. Images (Classes)} &
  \textbf{Gen./Imp. Pairs} \\ \hline
\multirow{3}{*}{\textbf{PolyU}}      & \textbf{5-fold} & \multirow{3}{*}{6,270 (418)} & 5,016 (418)        & -                & 1,254 (418)    & -                \\ \cline{2-2} \cline{4-7} 
                                     & \textbf{CW}     &                              & 4,180 (418)        & 18,810/8,715,300 & 2,090 (418)    & 4,180/2,178,825  \\ \cline{2-2} \cline{4-7} 
                                     & \textbf{OW}     &                              & 3,135 (209)        & 21,945/4,890,600 & 3,135 (209)    & 2,1945/4,890,600 \\ \hline
\multirow{3}{*}{\textbf{Cross-Eyed}} & \textbf{5-fold} & \multirow{3}{*}{1920 (240)}  & *1,680/1,440 (240) & -                & *240/480 (240) & -                \\ \cline{2-2} \cline{4-7} 
                                     & \textbf{CW}     &                              & 1,200 (240)        & 2,400/717,000    & 720 (240)      & 720/258,120      \\ \cline{2-2} \cline{4-7} 
                                     & \textbf{OW}     &                              & 960 (120)          & 3,360/456,960    & 960 (120)      & 3,360/456,960    \\ \hline
\end{tabular}
}
\caption{Summary of Train/Test partitions per database, spectrum, and protocol for the verification experiments, Both databases contain the same number of images in each spectrum. The Close-World (CW) and Open-World (OW) protocols with PolyU and Cross-Eyed are defined following \cite{depresion}. $^*$The values indicate the number of images when the test fold is composed of 1 or 2 images per class due to the number of images per user not being divisible by 5.}\label{tabla-gen-imp}
\end{table*}

\section{Methodology}
\label{Method}


We follow the pipeline shown in Fig. \ref{I20process} to combine the CST and CNNs architectures.
The $\sigma_1$ determines the frequency band in which CST acts, whereas $\sigma_2$  determines neighborhood size.
Preliminary results showed us that the highest frequencies in our datasets contained the most discriminative information (regarding identities in our study). The neighborhood size only influenced methods where alignment was an issue. 
We settled therefore for a value for $\sigma_1$ of $0.6$ to extract high-frequency
gradients and $\sigma_2$ of $4.0$ to define minimum neighborhood size. Furthermore, we set $\gamma$ to $0.1$ to reduce the magnitude difference between regions with subtle changes, such as skin tissue, and those with rapid changes, as introduced by zero-padding and masks. This helps to normalize the data.


We carried out biometric recognition experiments with six widely used CNN networks,  ResNet50, DenseNet121, VGG16, Xception, InceptionV3, MobileNetV2.

We used Tensorflow-Keras to download, initialize, train, and test the models. The CST process was transparent to the network and not affected by the training process. We use the standard models provided by the Keras library, only changing the last fully connected layer to match the number of users in the database and the input size to match the data type. We trained the networks using Stochastic Gradient Descent with a learning rate of $0.001$ and a momentum of $0.9$, except for VGG16, which did not use momentum and had a Clip-Value of $0.5$ due to convergence difficulties. The networks were trained for $100$ epochs with a batch size of $16$. No data augmentation was used during training. Model checkpoint was used to monitor the validation loss and recover the best-performing version after training. All training was conducted on a Windows 10 machine with an i7-8700, 32GB of RAM, and an Nvidia Titan V. 

Hyperparameters were deliberately kept standard across architectures to ensure that the performance gains isolate the structural benefit of the CST inputs, rather than reflecting optimized tuning.

\section{Databases, Metrics, and Protocol}
\label{DBMetricsProt}

\subsection{Databases}
\label{datasets}

We employed two commonly used periocular datasets in the experimentation: Cross-Eyed \cite{x-eyed2016}\cite{x-eyed2017}, and PolyU \cite{polyu}.

The \textbf{Cross-Eyed} dataset is a cross-spectral periocular database captured for the first Cross-Spectral Iris/Periocular Competition \cite{x-eyed2016}. Near-Infrared (NIR) and Visible (VIS) ocular images were captured simultaneously from $120$ subjects of different nationalities, ethnicities, and eye colors using a custom dual-spectrum image sensor under normal indoor illumination. For each of the $120$ subjects, $8$ images of both eyes were captured in both spectra, for a total of $3840$ images. They provide a version with the sclera masked for the periocular-only challenge, ensuring no iris information is used. We used the iris masks to normalize the images to have the same sclera radius, center, and orientation. Finally, they were zero-padded and cropped, so all have the same size and dimensions.

The \textbf{PolyU} database is a bi-spectral iris database captured using simultaneous bi-spectral imaging by the Hong Kong Polytechnic University. It provides NIR and VIS images captured simultaneously, with exact pixel correspondence between the spectral image versions. The database consists of $15$ instances per spectrum for both eyes of $209$ subjects, for a total of $12540$ iris images. Since the periocular region in this dataset is somewhat smaller, images are zero-padded to be squared while maintaining the aspect ratio.

\subsection{Metrics and Protocol}
\label{metricsprotocol}

To evaluate the performance of the proposed system, we carried out both biometric identification and verification experiments. For the identification case, due to limited training data, we used a 5-fold cross-validation strategy. We report the average accuracy over the five folds.

We also used a verification setting to compare our results with the SOA. In a verification scenario, we trained the networks first for biometric identification using standard softmax activation and cross-entropy loss. We then tested the network's verification performance by extracting the output feature vector of the second-to-last layer. Feature vectors from different images were compared using the Cosine similarity, following an all-against-all matching strategy.

For a fair comparison with the SOA under biometric verification, we followed the same protocol as in \cite{depresion}: the Close-World (CW) protocol, in which the training and test partition contains images of all users in the database without overlap, and the Open-World (OW) protocol, in which the users are split into training and testing along with all their images. For the Cross-Eyed dataset and the Close-World protocol, the first $5$ images of each user are used for training and the remaining $3$ for testing, while on the Open-World protocol, the first $120$ users, along with all their images, form the training partition and the last $120$ users the testing one. For PolyU and the Close-World setup, each user's first $10$ images are in the training set, and the last $5$ are in the test set. For the Open-World case, the first $209$ users are used for training and the last $209$ for testing. Table \ref{tabla-gen-imp} summarizes the number of images for each database, spectrum, and division.

\section{Results}\label{Discuss}

\subsection{Effect of CST input variables}




\begin{table}[h!]
\centering
\resizebox{0.90\textwidth}{!}{%
\begin{tabular}{c|c|c|c|c|c|c|c|c|c|c|} \cline{2-11}

      &\multicolumn{6}{|c|}{\textbf{Input}} &
  \multicolumn{2}{|c|}{\textbf{Cross-Eyed}} &
  \multicolumn{2}{|c|}{\textbf{PolyU}} \\ \cline{2-11}  
      &\multicolumn{1}{|c|}{\textbf{BW}} &
  \multicolumn{1}{|c|}{\textbf{$\left | I_{20} \right |$}} &
  \multicolumn{1}{|c|}{\textbf{$\angle I_{20}$}} &
  \multicolumn{1}{|c|}{\textbf{$\Re(I_{20})$}} &
  \multicolumn{1}{|c|}{\textbf{$\Im(I_{20})$}} &
  \textbf{$I_{11}$} &
  \multicolumn{1}{|c|}{\textbf{NIR}} &
  \textbf{VIS} &
  \multicolumn{1}{|c|}{\textbf{NIR}} &
  \textbf{VIS} \\ \hline  
   \multicolumn{1}{|c|}{\textbf{$1^{st}$}}   &\multicolumn{1}{|c|}{\textbf{X}} &
  \multicolumn{1}{|c|}{\textbf{}} &
  \multicolumn{1}{|c|}{\textbf{}} &
  \multicolumn{1}{|c|}{\textbf{}} &
  \multicolumn{1}{|c|}{\textbf{}} &
  \textbf{} &
  \multicolumn{1}{|c|}{97.8} &
  97.7 &
  \multicolumn{1}{|c|}{93.2} &
  94.5 \\ \hline  
  \multicolumn{1}{|c|}{\textbf{$2^{nd}$}}    &\multicolumn{1}{|c|}{\textbf{}} &
  \multicolumn{1}{|c|}{\textbf{X}} &
  \multicolumn{1}{|c|}{\textbf{}} &
  \multicolumn{1}{|c|}{\textbf{}} &
  \multicolumn{1}{|c|}{\textbf{}} &
  \textbf{} &
  \multicolumn{1}{|c|}{96.4} &
  97.1 &
  \multicolumn{1}{|c|}{93.7} &
  94.3 \\ \hline  
  \multicolumn{1}{|c|}{\textbf{$3^{rd}$}}    &\multicolumn{1}{|c|}{\textbf{}} &
  \multicolumn{1}{|c|}{\textbf{X}} &
  \multicolumn{1}{|c|}{\textbf{}} &
  \multicolumn{1}{|c|}{\textbf{}} &
  \multicolumn{1}{|c|}{\textbf{}} &
  \textbf{X} &
  \multicolumn{1}{|c|}{97.6} &
  97.6 &
  \multicolumn{1}{|c|}{94.7} &
  95.4 \\ \hline  
   \multicolumn{1}{|c|}{\textbf{$4^{th}$}}   &\multicolumn{1}{|c|}{\textbf{}} &
  \multicolumn{1}{|c|}{\textbf{X}} &
  \multicolumn{1}{|c|}{\textbf{X}} &
  \multicolumn{1}{|c|}{\textbf{}} &
  \multicolumn{1}{|c|}{\textbf{}} &
  \textbf{X} &
  \multicolumn{1}{|c|}{98.4} &
  98.4 &
  \multicolumn{1}{|c|}{95.6} &
  96.1 \\ \hline  
  \multicolumn{1}{|c|}{\textbf{$5^{th}$}}    &\multicolumn{1}{|c|}{\textbf{}} &
  \multicolumn{1}{|c|}{\textbf{}} &
  \multicolumn{1}{|c|}{\textbf{}} &
  \multicolumn{1}{|c|}{\textbf{X}} &
  \multicolumn{1}{|c|}{\textbf{X}} &
  \textbf{} &
  \multicolumn{1}{|c|}{98.2} &
  98.5 &
  \multicolumn{1}{|c|}{96.0} &
  96.0 \\ \hline  
  \multicolumn{1}{|c|}{\textbf{$6^{th}$}}    &\multicolumn{1}{|c|}{\textbf{}} &
  \multicolumn{1}{|c|}{\textbf{}} &
  \multicolumn{1}{|c|}{\textbf{}} &
  \multicolumn{1}{|c|}{\textbf{X}} &
  \multicolumn{1}{|c|}{\textbf{X}} &
  \textbf{X} &
  \multicolumn{1}{|c|}{98.5} &
  98.6 &
  \multicolumn{1}{|c|}{96.3} &
  \textbf{96.4} \\ \hline  
   \multicolumn{1}{|c|}{\textbf{$7^{th}$}}   &\multicolumn{1}{|c|}{\textbf{}} &
  \multicolumn{1}{|c|}{\textbf{X}} &
  \multicolumn{1}{|c|}{\textbf{}} &
  \multicolumn{1}{|c|}{\textbf{X}} &
  \multicolumn{1}{|c|}{\textbf{X}} &
  \textbf{X} &
  \multicolumn{1}{|c|}{98.4} &
  \textbf{98.7} &
  \multicolumn{1}{|c|}{\textbf{96.5}} &
  \textbf{96.4} \\ \hline  
   \multicolumn{1}{|c|}{\textbf{$8^{th}$}}   &\multicolumn{1}{|c|}{\textbf{X}} &
  \multicolumn{1}{|c|}{\textbf{}} &
  \multicolumn{1}{|c|}{\textbf{}} &
  \multicolumn{1}{|c|}{\textbf{X}} &
  \multicolumn{1}{|c|}{\textbf{X}} &
  \textbf{X} &
  \multicolumn{1}{|c|}{\textbf{98.6}} &
  98.2 &
  \multicolumn{1}{|c|}{\textbf{96.5}} &
  96.2 \\ \hline  
  \multicolumn{1}{|c|}{\textbf{$9^{th}$}}    &\multicolumn{1}{|c|}{\textbf{X}} &
  \multicolumn{1}{|c|}{\textbf{X}} &
  \multicolumn{1}{|c|}{\textbf{}} &
  \multicolumn{1}{|c|}{\textbf{X}} &
  \multicolumn{1}{|c|}{\textbf{X}} &
  \textbf{X} &
  \multicolumn{1}{|c|}{98.3} &
  98.3 &
  \multicolumn{1}{|c|}{\textbf{96.5}} &
  96.2 \\ \hline 
\end{tabular}
}
\caption{\label{tab:input_type} Average test accuracy from the 5-fold cross-validation identification experiments with respect to the variable type used at the input of a randomly initialized ResNet50.
}
\end{table}


\begin{figure*}[!h]
\centering
\includegraphics[width=0.90\textwidth]{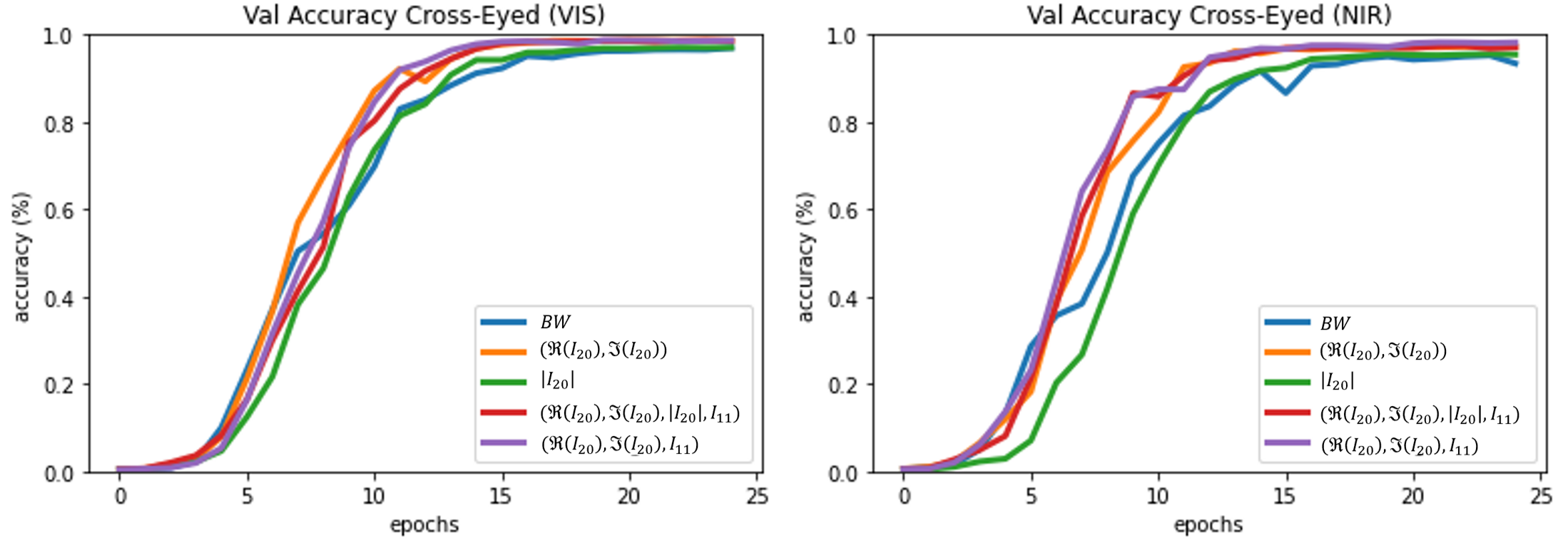}
\caption{Cross-Eyed database: accuracy during training for a randomly initialized ResNet50 network using different input data. \label{converge_cross-eyed}}
\end{figure*}

\begin{figure*}[!h]
\centering
\includegraphics[width=0.90\textwidth]{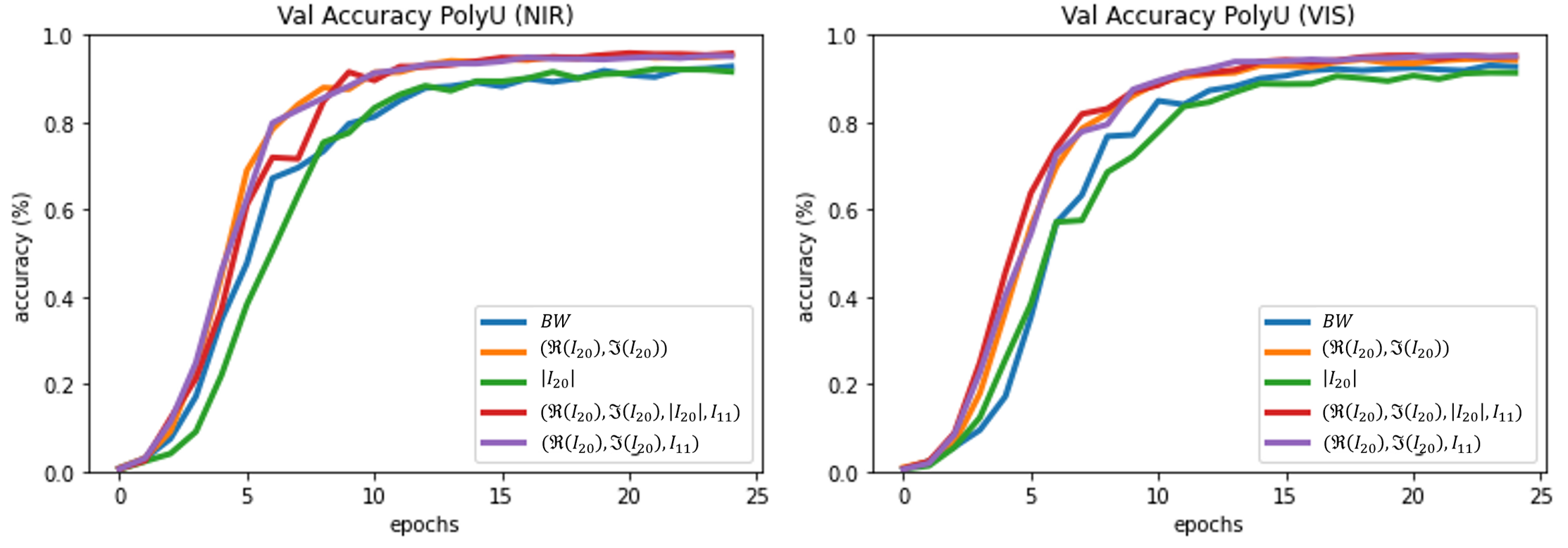}
\caption{PolyU database: accuracy during training for a randomly initialized ResNet50 network using different input data. \label{converge_polyu}}
\end{figure*}


The first step in our experimentation was to investigate how the representations of $I_{20}$ and $I_{11}$ affected the network's training and performance. Table \ref{tab:input_type} presents the identification accuracy results for these experiments. Additionally, in Figs. \ref{converge_cross-eyed} and \ref{converge_polyu}, we provide the validation accuracy during the training process for the Cross-Eyed and PolyU databases, respectively. 

Table \ref{tab:input_type} shows that using only $\left | I_{20} \right |$ achieves performance comparable to grayscale, and even better on PolyU and NIR. Adding $I_{11}$ increases average accuracy, matching grayscale on Cross-Eyed and improving results on PolyU, suggesting the upper fitting boundary is helpful. Including angle information further boosts accuracy across all settings, surpassing grayscale performance and indicating that the network handles angle discontinuities well. The consistent gains from adding magnitude, angle, and $I_{11}$ imply that each contributes useful information.

Using the real and imaginary parts of $I_{20}$ performs similarly to or better than polar features, and adding $I_{11}$ further improves performance. Explicitly adding $\left | I_{20} \right |$ yields a small additional gain. Combining grayscale with strong CST variants can help, though not always. Overall, the best results come from using $\Re(I_{20})$, $\Im(I_{20})$, $\left | I_{20} \right |$, and $I_{11}$, which jointly provide linear-pattern strength and continuous orientation cues. These trends hold across both datasets and spectra, indicating good generalization.

We then analyze the benefits of the proposed methods for network training (Figs. \ref{converge_polyu} and \ref{converge_cross-eyed}) for selected cases in Table \ref{tab:input_type}. Simply providing the real and imaginary parts (orange curves) already yields faster convergence and higher validation accuracy than using the grayscale version (blue). The best cases overall correspond to the addition of $I_{11}$ (purple) and $\left | I_{20} \right |$ (red) to the real and imaginary parts of $I_{20}$. Again, these benefits are observed regardless of the database and spectrum employed.

For the remainder of this paper, we retain the configuration of $\Re(I_{20})$,$\Im(I_{20})$, $I_{11}$ as input to the network (sixth row of Table \ref{tab:input_type}). This input provides the best balance between the system's accuracy and input size, with a standard number of channels for CNNs.

\subsection{Effect of network compression}
\label{compression}


\begin{figure*}[!h]
\centering
\includegraphics[width=0.90\textwidth]{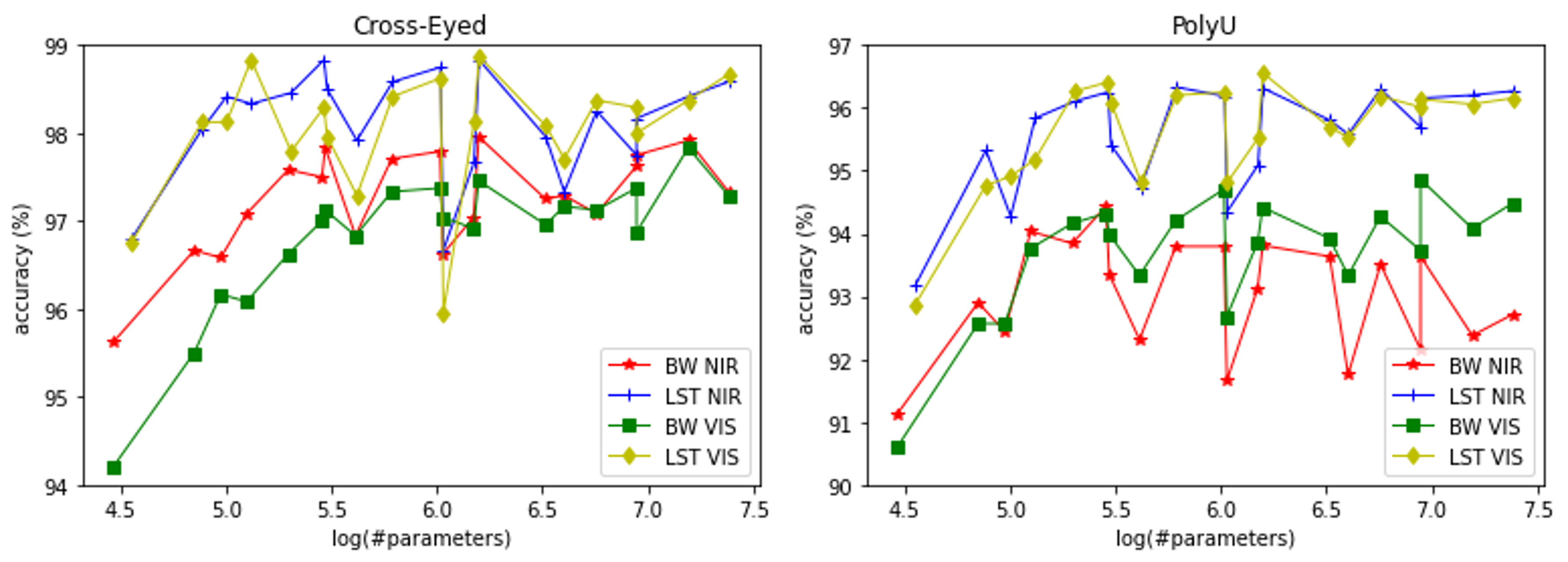}
\caption{Average test accuracy for different network compression levels. The number of parameters depends on the network width, controlled by the coefficient $\alpha$, and on the network depth, as explained in Section \ref{compression}. The complex data representation used at the input is $\Re(I_{20})$,$\Im(I_{20})$, $I_{11}$. 
\label{parametereffect}}
\end{figure*}


Our orientation-based inputs provide ready-to-use features, which may enable network compression and faster inference. We therefore varied the width, depth, and input resolution of ResNet50. Width (number of filters per layer) was scaled by a factor $\alpha \in \{0.2,0.4,0.6,0.8,1.0\}$ (with 1.0 as standard). The input resolutions used were 256x256, 128x128, 64x64, and 32x32. Depth was reduced at each resolution by truncating the network just before the next feature-map downsampling, preserving the abstraction level and output resolution.

Figure \ref{parametereffect} shows that accuracy correlates with the number of parameters, and the trends are consistent across datasets, spectra, and inputs. The structure-tensor input slightly increases the number of parameters compared to grayscale, due to additional input channels (an additional 6,272 parameters in the first layer for standard-width ResNet50), but this is negligible compared to the $\approx25.5$M parameters of ResNet50. Overall, the structure-tensor representation outperforms grayscale at nearly all model sizes, except one point in Cross-Eyed VIS setting at $\approx1$M parameters $(\log(\#parameters)=6)$, where maximum depth combined with minimum width $(\alpha=0.2)$ causes a sharp drop for all cases. Importantly, strong compression is possible with little loss in accuracy, and our input consistently yields better performance across compression levels, supporting smaller, faster models.

\begin{figure*}[!h]
\centering
\includegraphics[width=0.90\textwidth]{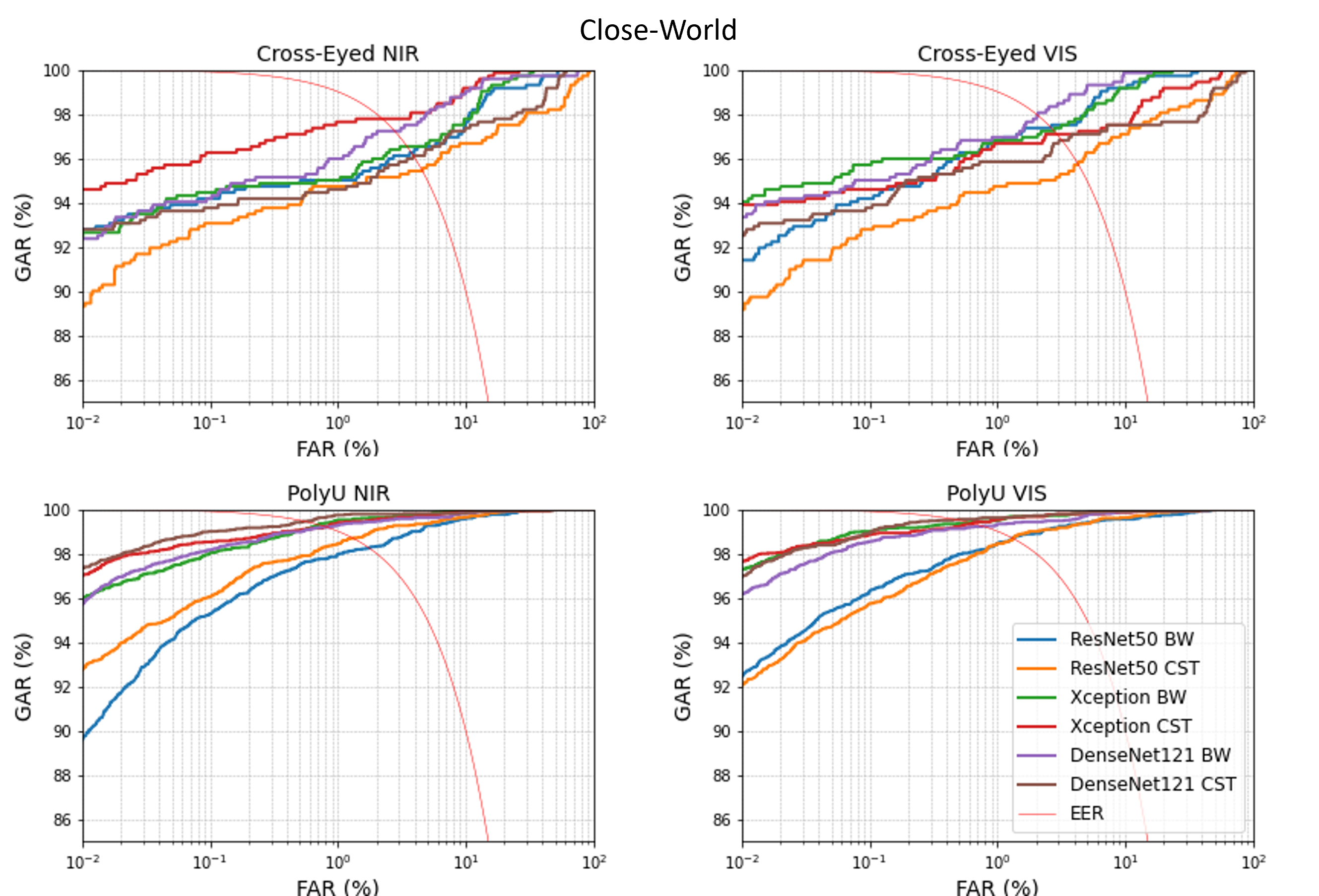}
\caption{ROC curves for the Close-World setting. The EER values and comparison with the SOA results are given in Table \ref{tab:SOA}
\label{ROC_CW}}
\end{figure*}

\begin{figure*}[!h]
\centering
\includegraphics[width=0.90\textwidth]{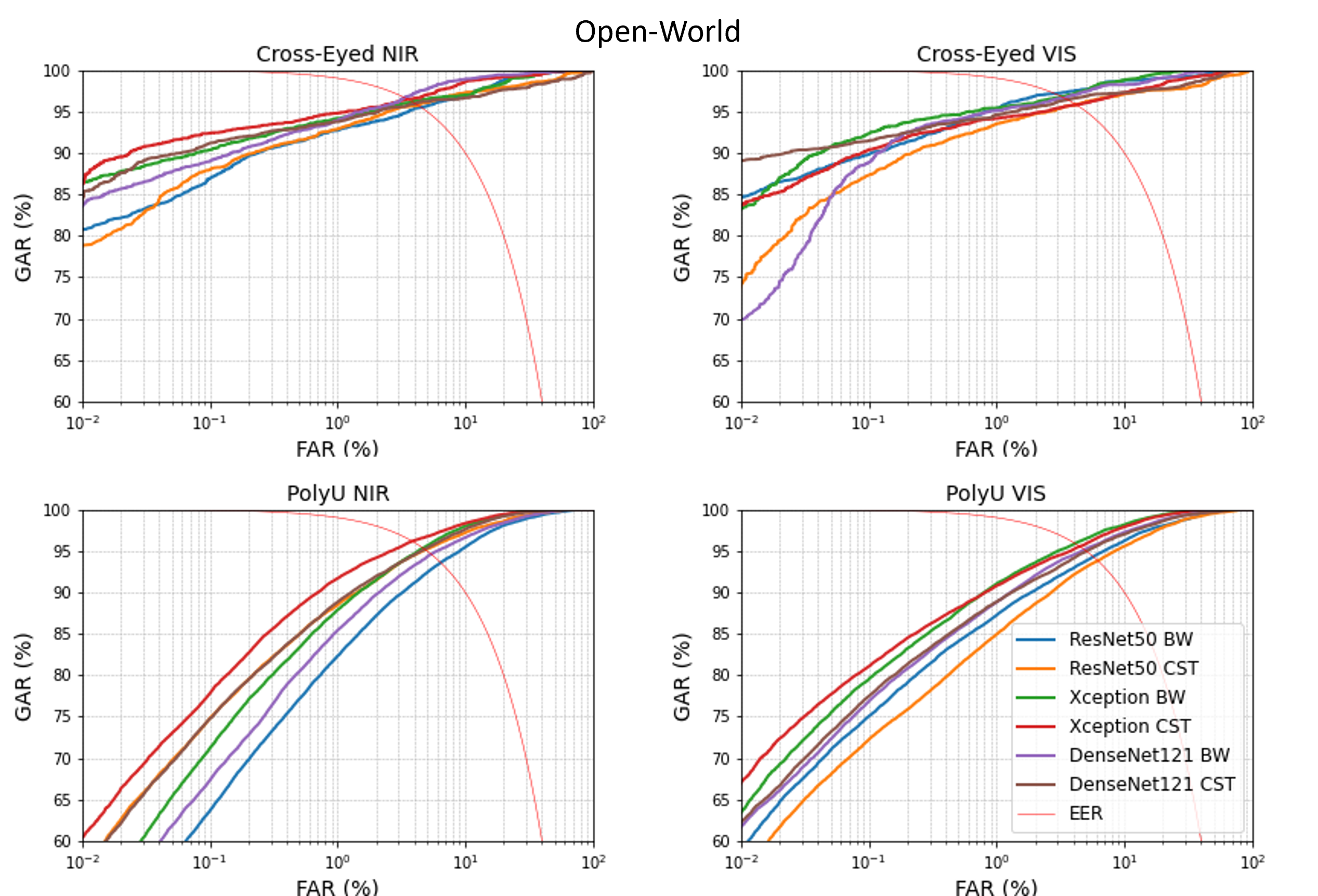}
\caption{ROC curves for the Open-World setting. The EER values and comparison with the SOA results are given in Table \ref{tab:SOA}
\label{ROC_OW}}
\end{figure*}

\subsection{Consistency across other CNN architectures}

\begin{table}[h!]
\centering
\resizebox{0.90\textwidth}{!}{%
\begin{tabular}{c|cccc|cccc|}
\cline{2-9}
 &
  \multicolumn{4}{c|}{\textbf{Cross-Eyed}} &
  \multicolumn{4}{c|}{\textbf{PolyU}} \\ \cline{2-9} 
 &
  \multicolumn{2}{c|}{\textbf{NIR}} &
  \multicolumn{2}{c|}{\textbf{VIS}} &
  \multicolumn{2}{c|}{\textbf{NIR}} &
  \multicolumn{2}{c|}{\textbf{VIS}} \\ \hline
\multicolumn{1}{|c|}{\textbf{Network}} &
  \multicolumn{1}{c|}{\textbf{BW}} &
  \multicolumn{1}{c|}{\textbf{CST}} &
  \multicolumn{1}{c|}{\textbf{BW}} &
  \textbf{CST} &
  \multicolumn{1}{c|}{\textbf{BW}} &
  \multicolumn{1}{c|}{\textbf{CST}} &
  \multicolumn{1}{c|}{\textbf{BW}} &
  \textbf{CST} \\ \hline
\multicolumn{1}{|c|}{\textbf{ResNet50}} &
  \multicolumn{1}{c|}{97.8} &
  \multicolumn{1}{c|}{\textbf{98.5}} &
  \multicolumn{1}{c|}{97.7} &
  \textbf{98.6} &
  \multicolumn{1}{c|}{93.3} &
  \multicolumn{1}{c|}{\textbf{96.3}} &
  \multicolumn{1}{c|}{94.5} &
  \textbf{96.4} \\ \hline
\multicolumn{1}{|c|}{\textbf{VGG16}} &
  \multicolumn{1}{c|}{95.4} &
  \multicolumn{1}{c|}{\textbf{99.0}} &
  \multicolumn{1}{c|}{96.5} &
  \textbf{98.8} &
  \multicolumn{1}{c|}{-} &
  \multicolumn{1}{c|}{-} &
  \multicolumn{1}{c|}{-} &
  - \\ \hline
\multicolumn{1}{|c|}{\textbf{DenseNet121}} &
  \multicolumn{1}{c|}{98.3} &
  \multicolumn{1}{c|}{\textbf{99.1}} &
  \multicolumn{1}{c|}{97.8} &
  \textbf{99.3} &
  \multicolumn{1}{c|}{95.8} &
  \multicolumn{1}{c|}{\cellcolor[HTML]{C0C0C0}\textbf{98.1}} &
  \multicolumn{1}{c|}{97.2} &
  \textbf{98.0} \\ \hline
\multicolumn{1}{|c|}{\textbf{Xception}} &
  \multicolumn{1}{c|}{98.5} &
  \multicolumn{1}{c|}{\cellcolor[HTML]{C0C0C0}\textbf{99.3}} &
  \multicolumn{1}{c|}{98.9} &
  \cellcolor[HTML]{C0C0C0}\textbf{99.6} &
  \multicolumn{1}{c|}{96.2} &
  \multicolumn{1}{c|}{\textbf{98.0}} &
  \multicolumn{1}{c|}{97.4} &
  \cellcolor[HTML]{C0C0C0}\textbf{98.1} \\ \hline
\multicolumn{1}{|c|}{\textbf{InceptionV3}} &
  \multicolumn{1}{c|}{97.8} &
  \multicolumn{1}{c|}{\textbf{98.9}} &
  \multicolumn{1}{c|}{97.9} &
  \textbf{98.3} &
  \multicolumn{1}{c|}{93.8} &
  \multicolumn{1}{c|}{\textbf{96.5}} &
  \multicolumn{1}{c|}{94.9} &
  \textbf{95.9} \\ \hline
\multicolumn{1}{|c|}{\textbf{MobileNetV2}} &
  \multicolumn{1}{c|}{98.6} &
  \multicolumn{1}{c|}{\textbf{99.0}} &
  \multicolumn{1}{c|}{98.4} &
  \textbf{99.2} &
  \multicolumn{1}{c|}{94.5} &
  \multicolumn{1}{c|}{\textbf{97.3}} &
  \multicolumn{1}{c|}{96.1} &
  \textbf{97.4} \\ \hline
\end{tabular}
}
\caption{\label{tab:best_network} Average test accuracy for each network. The complex data representation used at the input is $\Re(I_{20})$,$\Im(I_{20})$, $I_{11}$. The bold numbers indicate the best case among BW (grayscale images) and CST (our approach). The gray cells indicate the best result per column. 
}
\end{table}


We also tested the method on other CNN architectures (Table \ref{tab:best_network}) and found consistent gains: the complex structure tensor input outperforms grayscale for every network and database. On Cross-Eyed, VGG16 improves by $3.6\%$ (NIR) and $2.3\%$ (VIS). On PolyU, ResNet50 improves by $3.3\%$ (NIR) and $1.9\%$ (VIS), with additional gains for InceptionV3 and MobileNetV2 on PolyU-NIR ($2.7\%$ and $2.8\%$). Xception performs best overall for both inputs, reaching up to 99.6\% on Cross-Eyed VIS and about $98\%–99\%$ elsewhere. These results indicate the method generalizes well across architectures.

\subsection{Comparison with previous studies}


Table \ref{tab:SOA} compares our results with prior work on the same databases \cite{hernandez2023one}\cite{depresion}. We follow the same biometric verification setup under the CW/OW protocols of Table \ref{tabla-gen-imp} (see Section \ref{metricsprotocol}), reporting test-set EER and showing full performance via the ROC curves in Figs. \ref{ROC_CW} and \ref{ROC_OW}. For these experiments, CST uses the input set BW, $\Re(I_{20})$, $\Im(I_{20})$, and $I_{11}$, combining grayscale with orientation-rich cues.

\begin{table}[h!]
\centering
\resizebox{0.90\textwidth}{!}{%
\begin{tabular}{cc|cccc|cccc|}
\cline{3-10}
  &
    &
   \multicolumn{2}{c|}{\textbf{Cross-Eyed (CW)}} &
   \multicolumn{2}{c|}{\textbf{PolyU (CW)}} &
   \multicolumn{2}{c|}{\textbf{Cross-Eyed (OW)}} &
   \multicolumn{2}{c|}{\textbf{PolyU (OW)}} \\ \cline{3-10}
  &
    &
   \multicolumn{1}{c|}{\textbf{NIR}} &
   \multicolumn{1}{c|}{\textbf{VIS}} &
   \multicolumn{1}{c|}{\textbf{NIR}} &
   \textbf{VIS} &
   \multicolumn{1}{c|}{\textbf{NIR}} &
   \multicolumn{1}{c|}{\textbf{VIS}} &
   \multicolumn{1}{c|}{\textbf{NIR}} &
   \textbf{VIS} \\ \hline
\multicolumn{1}{|c|}{\multirow{4}{*}{\textbf{ResNet50}}} &
   \textbf{BW} &
   \multicolumn{1}{c|}{3.7} &
   \multicolumn{1}{c|}{2.6} &
   \multicolumn{1}{c|}{1.8} &
   1.3 &
   \multicolumn{1}{c|}{4.5} &
   \multicolumn{1}{c|}{\textbf{2.9}} &
   \multicolumn{1}{c|}{6.4} &
   5.7 \\ \cline{2-10}
\multicolumn{1}{|c|}{} &
   \textbf{CST} &
   \multicolumn{1}{c|}{4.6} &
   \multicolumn{1}{c|}{4.3} &
   \multicolumn{1}{c|}{\cellcolor[HTML]{C0C0C0}1.3} &
   \cellcolor[HTML]{C0C0C0}1.3 &
   \multicolumn{1}{c|}{\cellcolor[HTML]{C0C0C0}4.3} &
   \multicolumn{1}{c|}{4.2} &
   \multicolumn{1}{c|}{\cellcolor[HTML]{C0C0C0}4.9} &
   6.2 \\ \cline{2-10}
\multicolumn{1}{|c|}{} &
   \cite{depresion} &
   \multicolumn{1}{c|}{\textbf{1.8}} &
   \multicolumn{1}{c|}{1.7} &
   \multicolumn{1}{c|}{0.68} &
   0.61 &
   \multicolumn{1}{c|}{3.5} &
   \multicolumn{1}{c|}{3.4} &
   \multicolumn{1}{c|}{4.00} &
   \textbf{3.94} \\ \cline{2-10}
\multicolumn{1}{|c|}{} &
   \cite{hernandez2023one} &
   \multicolumn{1}{c|}{-} &
   \multicolumn{1}{c|}{\textbf{1.3}} &
   \multicolumn{1}{c|}{-} &
   3.4 &
   \multicolumn{1}{c|}{-} &
   \multicolumn{1}{c|}{\textbf{1.2}} &
   \multicolumn{1}{c|}{-} &
   7.76 \\ \hline
\multicolumn{1}{|c|}{\multirow{2}{*}{\textbf{Xception}}} &
   \textbf{BW} &
   \multicolumn{1}{c|}{3.5} &
   \multicolumn{1}{c|}{2.8} &
   \multicolumn{1}{c|}{0.7} &
   0.6 &
   \multicolumn{1}{c|}{3.8} &
   \multicolumn{1}{c|}{3.2} &
   \multicolumn{1}{c|}{4.7} &
   \textbf{4.1} \\ \cline{2-10}
\multicolumn{1}{|c|}{} &
   \textbf{CST} &
   \multicolumn{1}{c|}{\cellcolor[HTML]{C0C0C0}\textbf{2.2}} &
   \multicolumn{1}{c|}{2.9} &
   \multicolumn{1}{c|}{0.8} &
   \cellcolor[HTML]{C0C0C0}0.6 &
   \multicolumn{1}{c|}{\cellcolor[HTML]{C0C0C0}3.5} &
   \multicolumn{1}{c|}{4.2} &
   \multicolumn{1}{c|}{\cellcolor[HTML]{C0C0C0}\textbf{3.8}} &
   4.5 \\ \hline
\multicolumn{1}{|c|}{\multirow{3}{*}{\textbf{DenseNet121}}} &
   \textbf{BW} &
   \multicolumn{1}{c|}{2.8} &
   \multicolumn{1}{c|}{\textbf{2.1}} &
   \multicolumn{1}{c|}{0.8} &
   0.8 &
   \multicolumn{1}{c|}{\textbf{3.3}} &
   \multicolumn{1}{c|}{3.4} &
   \multicolumn{1}{c|}{5.4} &
   4.9 \\ \cline{2-10}
\multicolumn{1}{|c|}{} &
   \textbf{CST} &
   \multicolumn{1}{c|}{3.9} &
   \multicolumn{1}{c|}{3.2} &
   \multicolumn{1}{c|}{\cellcolor[HTML]{C0C0C0}\textbf{0.5}} &
   \cellcolor[HTML]{C0C0C0}\textbf{0.5} &
   \multicolumn{1}{c|}{4.1} &
   \multicolumn{1}{c|}{3.6} &
   \multicolumn{1}{c|}{\cellcolor[HTML]{C0C0C0}4.8} &
   5.0 \\ \cline{2-10}
\multicolumn{1}{|c|}{} &
   \cite{hernandez2023one} &
   \multicolumn{1}{c|}{-} &
   \multicolumn{1}{c|}{1.6} &
   \multicolumn{1}{c|}{-} &
   2.49 &
   \multicolumn{1}{c|}{-} &
   \multicolumn{1}{c|}{1.7} &
   \multicolumn{1}{c|}{-} &
   5.82 \\ \hline
\end{tabular}
}
\caption{\label{tab:SOA}
CNN networks comparison with absence and presence of CST input, as well as SOA results for each database and spectrum. Bold cells show the best-overall and our best result for each particular setting (column). Shaded cells emphasize improvement with the presence of the CST over standard networks.
}
\end{table}


Results are more mixed than in earlier sections: CST does not always outperform grayscale, especially on Cross-Eyed. We attribute this partly to evaluating a single fixed test split (rather than 5-fold cross-validation) and to Cross-Eyed’s limited training data, since CST increases the number of input channels. On PolyU, where more training data is available, CST matches or improves on grayscale in most cases, suggesting that verification benefits are more apparent with sufficient data.


The ROC curves (Figs. \ref{ROC_CW} and \ref{ROC_OW}) show that CST can still be preferable at low FAR, where high genuine acceptance is required, for example with Xception on PolyU (NIR, CW) and on PolyU (VIS, OW). In the OW setting, CST generally performs better at low FAR, with negligible differences on PolyU-VIS; for DenseNet121 on Cross-Eyed NIR (CW), the gap narrows at low FAR as well. Overall, the ROC behavior supports earlier identification results: adding texture-orientation information can improve recognition when models are trained from scratch.





Compared to published studies \cite{hernandez2023one}\cite{depresion}, CST achieves state-of-the-art-level results on PolyU with Xception/DenseNet121 in CW (both spectra) and in OW for PolyU-NIR, despite no pre-training. For ResNet50, CST comes within about 1 percentage point of the best reported results on PolyU (CW and OW, NIR), and in other cases trails by roughly 2–3 points. Since \cite{depresion} relies on face-recognition pre-training (e.g., VGGFace2 and MS1M), our grayscale+orientation input strategy may benefit further from similar pre-training on large-scale modalities. Finally, \cite{hernandez2023one} improved verification by selecting informative network layers, which is a complementary direction to explore alongside CST.

\subsection{Empirical Evidence for Explainability and Representation Dynamics}

The consistent performance improvements observed with Complex Structure Tensor (CST) inputs provide empirical evidence regarding the representational limits of standard CNNs. In conventional pixel representations, a CNN must implicitly learn to approximate optimal, Gabor-like orientation filters in its earliest layers. Our results suggest that relying on this implicit learning leads to inefficient use of parameters and opaque feature extraction. By utilizing CST inputs ($\mathfrak{R}(I_{20})$, $\mathfrak{S}(I_{20})$, $I_{11}$), we replace this black-box approximation with a deterministic, structured orientation prior. This enhances the explainability of the network's decision-making process in three concrete ways: 

\begin{enumerate}
    \item Translation Invariance: The network is explicitly fed features that are translation-invariant within a given texture pattern, removing the burden on the CNN of learning this invariance through data augmentation or parameter bloat.
    \item Explicit Certainty Measures: The inclusion of $I_{11}$ provides the network with a tight upper bound for the confidence of the linear fit. Unlike raw grayscale pixels, the network receives a direct, mathematically grounded signal indicating the certainty and strength of a structural pattern.
    \item Parameter Efficiency: The network compression results (Section \ref{compression}) demonstrate that when explicit orientation features are provided, network width and depth can be drastically reduced with minimal accuracy loss. This provides evidence that standard CNNs allocate significant parameter space solely to resolving low-level orientation ambiguities.
\end{enumerate}
Ultimately, these results demonstrate that upfront, biologically inspired feature extraction does not merely boost benchmark scores; it fundamentally shifts the network's optimization landscape, forcing it to base its classifications on verifiable, structured geometric patterns.

\section{Conclusions}\label{Conclusions}


This study evaluates Complex Structure Tensor (CST) features as an orientation-rich texture information as input to CNNs. Instead of feeding grayscale images, we preprocess them to emphasize local orientations using second-order complex moments, yielding a complex representation that encodes the magnitude and angle of dominant linear patterns. We validate the approach on periocular recognition using six CNN architectures and the Cross-Eyed and PolyU datasets, covering both NIR and VIS spectra.


Our method extracts single-band texture components and averages them over neighborhoods to reduce noise (controlled by $\sigma_1$ and $\sigma_2$). Although this reduces the input feature space, results show that the retained texture information is sufficient, and often superior, for improving identification accuracy and accelerating convergence across datasets, spectra, and architectures. This provides empirical evidence confirming our hypothesis that standard CNNs struggle to reliably extract these cues from grayscale images, especially with limited data.


Because CST provides more “ready-made” low-level features, we also explore reducing network depth and width, finding that larger compression is possible while maintaining better performance. In verification, CST improves results when training data is sufficient (notably on PolyU), while gains are less consistent on Cross-Eyed, likely due to limited training data and evaluation on a single fixed test split.


Providing explicit orientation features also improves interpretability: networks can base decisions on structured, translation-invariant texture cues with well-defined properties (e.g., unbiased direction and certainty measures), reducing the burden of learning them implicitly. This aligns with work that replaces early CNN layers with (learnable) Gabor filters, which similarly improve accuracy and convergence. Our results reinforce the idea that such biologically inspired low-level cues can be hard for CNNs to access unless made explicit and remain fairly underexplored.


Next, we plan to make CST filter hyperparameters learnable and extend the analysis on their effects and influence on the network's performance. Since symmetry filters are defined by a few interpretable parameters, learning these could speed training and further improve interpretability and efficiency. We also aim to combine filter families to target specific symmetry patterns (as in \cite{symmetryfingerprint}) alongside texture, potentially improving detection and segmentation of structures such as the iris region.


%
%
\bibliographystyle{splncs04}
\bibliography{main}
\end{document}